\pdfoutput=1

\documentclass[11pt]{article}
\usepackage[table,xcdraw,dvipsnames]{xcolor}
\usepackage{colortbl}          
\usepackage{booktabs}          
\usepackage{caption}           
\usepackage{tcolorbox}
\usepackage{enumitem}
\usepackage{stfloats}
\usepackage[preprint]{acl}

\usepackage{times}
\usepackage{latexsym}
\usepackage{amsmath}
\usepackage[T1]{fontenc}

\usepackage[utf8]{inputenc}

\usepackage{microtype}

\usepackage{inconsolata}

\usepackage{graphicx}
\makeatletter
 \twocolumn 

\def\maketitle{\par
 \begingroup
   \def\thefootnote{\fnsymbol{footnote}}
   \def\@makefnmark{\hbox to 0pt{$^{\@thefnmark}$\hss}}
   \twocolumn[\@maketitle] \@thanks
 \endgroup
 \setcounter{footnote}{0}
 \let\maketitle\relax \let\@maketitle\relax
 \gdef\@thanks{}\gdef\@author{}\gdef\@title{}\let\thanks\relax}
\def\@maketitle{\vbox to \titlebox{\hsize\textwidth
 \linewidth\hsize \vskip 0.125in minus 0.125in \centering
 {\Large\bfseries \@title \par} \vskip 0.2in plus 1fil minus 0.1in
 {\def\and{\unskip\enspace{\rmfamily and}\enspace}%
  \def\And{\end{tabular}\hss \egroup \hskip 1in plus 2fil
           \hbox to 0pt\bgroup\hss \begin{tabular}[t]{c}\bfseries}%
  \def\AND{\end{tabular}\hss\egroup \hfil\hfil\egroup
          \vskip 0.25in plus 1fil minus 0.125in
           \hbox to \linewidth\bgroup\large \hfil\hfil
             \hbox to 0pt\bgroup\hss \begin{tabular}[t]{c}\bfseries}
  \hbox to \linewidth\bgroup\large \hfil\hfil
    \hbox to 0pt\bgroup\hss \begin{tabular}[t]{c}\bfseries\@author
 \end{tabular}\hss\egroup
    \hfil\hfil\egroup}
  \vskip 0.3in plus 2fil minus 0.1in
}}
\makeatother
\title{TUMLU: A Unified and Native Language Understanding Benchmark for Turkic Languages}

\author{
 \textbf{Jafar Isbarov\textsuperscript{\textcolor{blue}{$\alpha$}1,2,3}},
 \textbf{Arofat Akhundjanova\textsuperscript{4}},
 \textbf{Mammad Hajili\textsuperscript{5}},
 \textbf{Kavsar Huseynova\textsuperscript{6}},
\\
 \textbf{Dmitry Gaynullin\textsuperscript{15}},
 \textbf{Anar Rzayev\textsuperscript{7}},
 \textbf{Osman Tursun\textsuperscript{8}},
  \textbf{Aizirek Turdubaeva\textsuperscript{7}},
 \textbf{Ilshat Saetov\textsuperscript{9}},\\
 \textbf{Rinat Kharisov\textsuperscript{15}},
 \textbf{Saule Belginova\textsuperscript{10}},
 \textbf{Ariana Kenbayeva\textsuperscript{11}},
 \textbf{Amina Alisheva\textsuperscript{11}},\\
 \textbf{Abdullatif Köksal\textsuperscript{\textcolor{blue}{$\beta$}12,13}},
 \textbf{Samir Rustamov\textsuperscript{\textcolor{blue}{$\beta$}14}},
 \textbf{Duygu Ataman\textsuperscript{\textcolor{blue}{$\beta$}2}}
\\
\\
 \textsuperscript{1}eiLink R\&D Center, Khazar University
 \textsuperscript{2}New York University,
 \textsuperscript{3}The George Washington University,\\
 \textsuperscript{4}Saarland University,
 \textsuperscript{5}Microsoft,
 \textsuperscript{6}Baku Higher Oil School,
 \textsuperscript{7}KAIST,\\
 \textsuperscript{8}Queensland University of Technology,
 \textsuperscript{9}CETOBaC EHESS,
 \textsuperscript{10}Turan University,\\
 \textsuperscript{11}Nazarbayev University
 \textsuperscript{12}LMU Munich,
\textsuperscript{13}MCML,
\textsuperscript{14}ADA University,
 \textsuperscript{15}Independent researcher
\\
\\
\small{\textsuperscript{\textcolor{blue}{$\alpha$}}First author. \textsuperscript{\textcolor{blue}{$\beta$}}Advisors.}\\
 \small{
   \textbf{Correspondence:} \href{mailto:jafar.isbarov@gwu}{jafar.isbarov@gwu.edu}
 }}

\begin{document}
\maketitle
\begin{abstract}
Being able to thoroughly assess massive multi-task language understanding (MMLU) capabilities is essential for advancing the applicability of multilingual language models. However, preparing such benchmarks in high quality native language is often costly and therefore limits the representativeness of evaluation datasets. While recent efforts focused on building more inclusive MMLU benchmarks, these are conventionally built using machine translation from high-resource languages, which may introduce errors and fail to account for the linguistic and cultural intricacies of the target languages. In this paper, we address the lack of native language MMLU benchmark especially in the under-represented Turkic language family with distinct morphosyntactic and cultural characteristics. We propose two benchmarks for Turkic language MMLU: TUMLU is a comprehensive, multilingual, and natively developed language understanding benchmark specifically designed for Turkic languages. It consists of middle- and high-school level questions spanning 11 academic subjects in Azerbaijani, Crimean Tatar, Karakalpak, Kazakh, Kyrgyz, Tatar, Turkish, Uyghur, and Uzbek. We also present TUMLU-mini, a more concise, balanced, and manually verified subset of the dataset. Using this dataset, we systematically evaluate a diverse range of open and proprietary multilingual large language models (LLMs), including Claude, Gemini, GPT, and LLaMA, offering an in-depth analysis of their performance across different languages, subjects, and alphabets. To promote further research and development in multilingual language understanding, we release TUMLU-mini and all corresponding evaluation scripts\footnote{\url{https://github.com/ceferisbarov/TUMLU}}.

\end{abstract}

\section{Introduction}
\begin{figure*}[t]
    \centering
    \includegraphics[width=\linewidth]{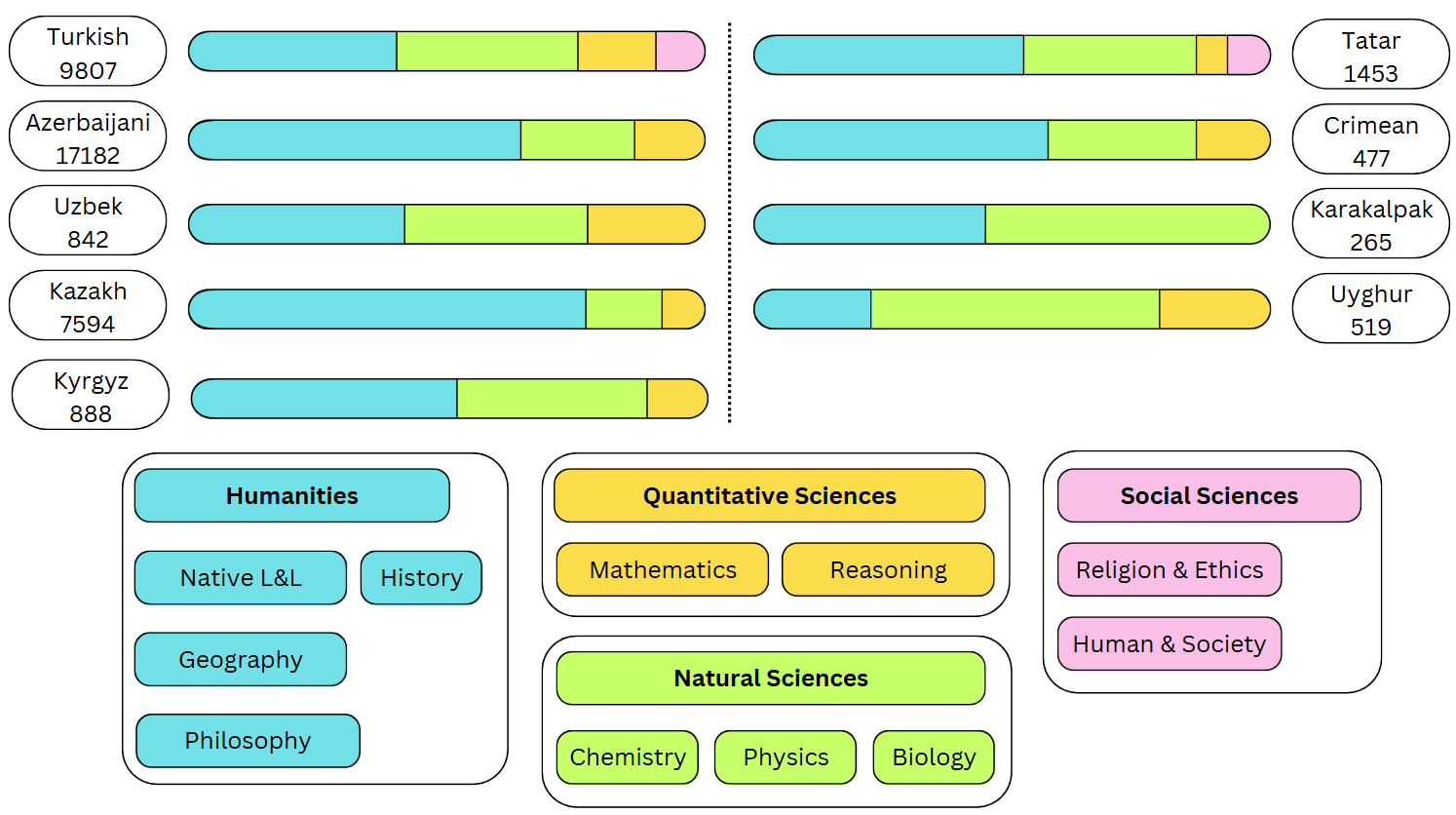}
    \caption{Distribution of subjects across languages in TUMLU. Numbers next to language names indicate the total question count. Left: middle- and high-resource languages; Right: low-resource languages.}
    \label{fig:num_samples}
\end{figure*}

Language understanding encompasses a system's ability to interpret and derive meaning from human language, incorporating syntax, semantics, and context. Evaluating language models hinges on this capability, as it ensures coherence, contextual relevance, and accuracy. Benchmarking is integral to assessing these models, particularly with the rapid advancements in Large Language Models (LLMs), which now support multiple languages \cite{qwen25,gemma2,llama3} and excel in complex reasoning tasks such as mathematical, scientific, and coding-related inquiries \cite{hurst2024gpt,claude35,gemini15,llama3}. However, the scarcity of robust natural language understanding (NLU) benchmarks capturing diverse linguistic and cultural contexts remains a challenge. Notably, LLM performance declines in low-resource languages, which are often underrepresented in training data, highlighting the need for more inclusive evaluation frameworks.

The majority of benchmarks included in top leaderboards where cutting-edge LLMs are evaluated are majorly prepared in English \cite{mmlu,suzgun2022challenging,mmlupro,superglue}. In order to extend the applicability of LLM evaluation in more languages, 
recent efforts were undertaken to build more multilingual NLU benchmarks \cite{lai2023okapi}, however, most of these either cover a limited set of high-resourced languages, or the multilingual examples are generated by translating original examples from Western-centric languages, thus failing to capture cultural nuances inherent in different languages. Due to the multi-dimensional nature of the reasoning task, language-specific benchmarks especially when translated into other languages also fail to represent the actual usage as well as demonstrating reasoning in the native language., and may further introduce issues such as translationese \cite{translationese} and cultural misalignment \cite{include}. On the other end of the spectrum, there are efforts to bridge that gap for a particular language, for example, African languages \cite{bayes2024uhura}, Arabic \cite{arabicmmlu}, Chinese \cite{cmmlu}, and Turkish \cite{turkishmmlu}.

In this paper, we focus on building a truly representative and inclusive single-language family benchmark to address previous problems and provide a challenging setting for LLM evaluation. TUMLU (Turkic Unified Multilingual Language Understanding) benchmark covers the following languages: Azerbaijani, Crimean Tatar, Turkish, Uyghur, Uzbek, Karakalpak, Kyrgyz, Kazakh, and Tatar.
The dataset consists of 4-choice questions at middle- and high-school levels. It consists of 38139 questions across 8 languages and 11 subjects (see Figure \ref{fig:num_samples} for a higher-level breakdown across languages). It is the first such benchmark to include Uyghur, Karakalpak, Tatar, or Crimean Tatar. It is also a significant improvement over existing benchmarks for Azerbaijani, Kazakh, Kyrgyz, and Uzbek. The Turkish dataset is TurkishMMLU, which was a separate project \cite{turkishmmlu}. 
The benchmark is also representative in terms of different scripts by including questions and answers in chosen languages in Latin, Cyrillic, and Arabic scripts. These datasets are transliterated such that it could be possible to get a dual dataset with the same content in two scripts for further comparative studies. We use these dual datasets to compare the performance of LLMs across different scripts.

We also release a more balanced and manually verified version of the dataset called TUMLU-mini, which contains 100 questions per subject (unless there are less than 100 for the said subject in a particular language). We use this version to test SOTA open-source and proprietary models of various sizes. We evaluated them in two settings: few-shot and chain-of-thought (CoT) reasoning \cite{cot}. Our initial results show that proprietary models remain the best option for Turkic languages.

\section{Related Work}
\paragraph{Language understanding benchmarks}
Multi-task language understanding evaluation benchmarks play an important role in the evaluation of LLMs. Early benchmarks concentrated on general natural language understanding. GLUE \cite{glue} and SuperGLUE \cite{superglue} were two such benchmarks that were widely adopted by the research community. These benchmarks were saturated quickly, due to the development of better LLMs. However, LLMs struggled more against benchmarks that required knowledge and reasoning. MMLU \cite{mmlu} and MMLU-Pro \cite{mmlupro} were more challenging since they required not only language understanding but also world knowledge. These general-purpose benchmarks gradually gave way to higher-level and more specialized benchmarks such as MATH \cite{math}, GPQA \cite{gpqa}, and MUSR \cite{musr}.

\paragraph{Multilingual benchmarks}
The development of multilingual LLMs also necessitated challenging multilingual benchmarks. Most of these benchmarks were developed through machine translation \cite{xnli, globalmmlu}. However, such datasets have been shown to contain cultural biases and translation artifacts \cite{translationese}. Global MMLU relied on machine and professional translation to \cite{globalmmlu}. INCLUDE consists of native data \cite{include}, but it is imbalanced, with different subject distributions in different languages. There is also a significant difference in required knowledge levels between languages, making a direct comparison impossible.

\paragraph{Benchmarks for Turkic languages}
SeaEval was one of the first LLM benchmarks to include Turkish \cite{seaeval}. Global MMLU contains Kyrgyz and Turkish subsets. INCLUDE contains Azerbaijani and Kazakh. MRL 2024 Shared Task on Multi-lingual Multi-task Information Retrieval \cite{mrl2024} contains an Azerbaijani dataset, but it contains general language understanding tasks rather than world knowledge. Kardeş-NLU has introduced a multilingual language understanding benchmark \cite{kardes}. But again, this benchmark contains general language understanding tasks that require no world knowledge. There are also monolingual benchmarks. Mukayese was one of the earliest general language understanding benchmarks in Turkish \cite{mukayese}. TurkishMMLU and TR-MMLU \cite{tr-mmlu} were the first native MMLU alternatives for the Turkish language. Another pilot study was performed to evaluate LLMs in Kazakh language \cite{dollmsspeakkazakh}. KazMMLU is another monolingual MMLU-style benchmark for the Kazakh language \cite{kazmmlu}. While there are no peer-reviewed monolingual MMLU alternatives for Azerbaijani, there is a general language understanding benchmark \cite{allma}.

\begin{figure}
    \centering
    \includegraphics[width=0.9\linewidth]{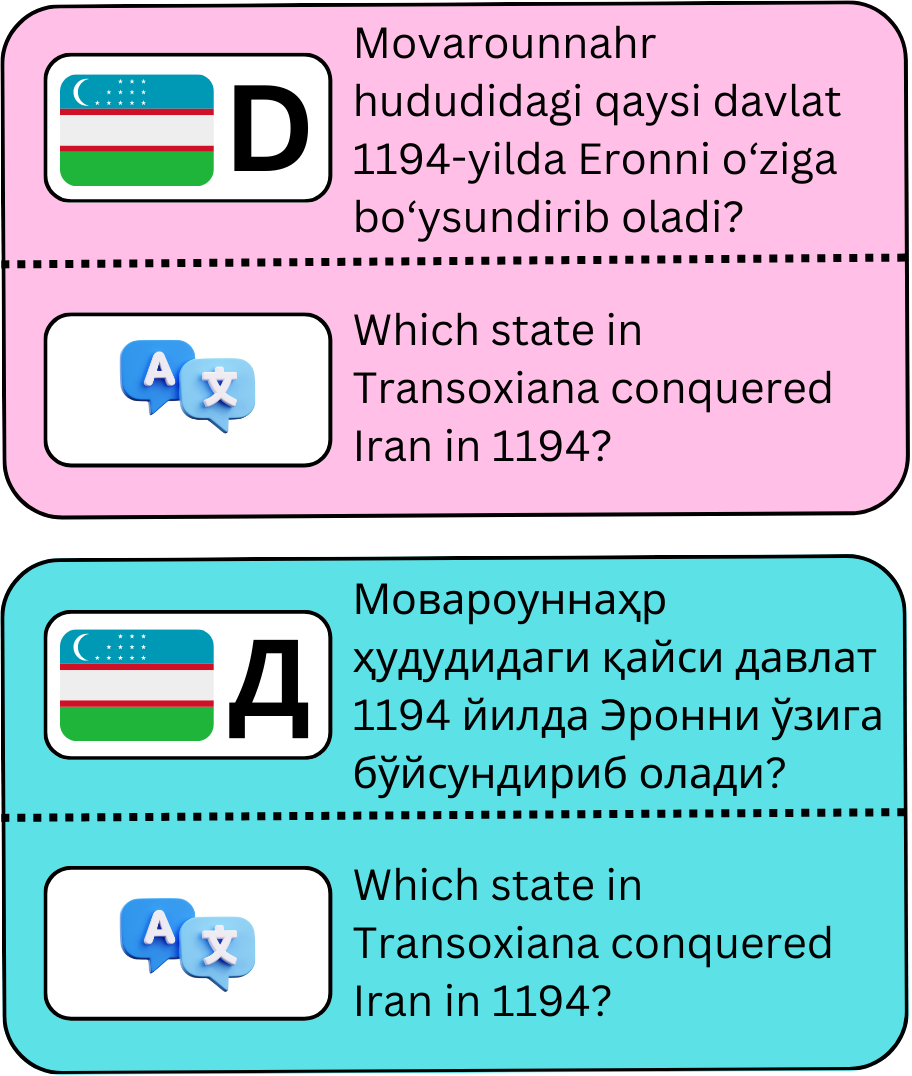}
    \caption{A sample question from the parallel Uzbek dataset, available in both Cyrillic and Latin alphabets. This enables comparison of LLM performance across different scripts. English translation is provided for clarity.}
    \label{fig:dual_dataset}
\end{figure}

\section{TUMLU}
TUMLU is a multilingual and multitask dataset containing 38139 multiple-choice questions across 8 languages and 11 subjects. All questions are at middle or high school level. The majority are sample or official questions for university entrance exams of respective countries.
\paragraph{Data collection} Data was collected from publicly available books and websites. In original form, questions had 2 to 5 choices. In cases where more than 4 choices were available, we removed an incorrect choice. If less than 4 choices were available, we left the question as-is. Except for Language and Literature questions in Crimean Tatar, all questions have 4 choices in the final version. 

After collecting the data, native speakers of each language manually verified the quality of a random sample from each subject. In languages such as Azerbaijani where questions were developed by the community, around 10 \% of the questions were either invalid or had incorrect answers.

We also created 5 CoT prompts per subject in Azerbaijani, Kazakh, Turkish, and Uzbek. Azerbaijani, Kazakh, and Uzbek prompts were created manually by native speakers. Turkish prompts were adapted from the TurkishMMLU project by changing the number of choices from 5 to 4. These prompts allowed us to compare the no-CoT and CoT performance of models. We did not create CoT prompts for other languages, because we did not have native speakers to validate their quality. CoT samples in Azerbaijani are available in Appendix \ref{sec:prompt-samples}.

\begin{table}[t]
    \centering
    \begin{tabular}{lrr}
    \hline
    \textbf{Language} & \textbf{Question} & \textbf{Answer} \\
    \hline
    Azerbaijani & 63.1 & 28.0 \\
    Crimean Tatar & 113.5 & 67.3 \\
    Karakalpak & 112.3 & 65.3 \\
    Kazakh & 96.8 & 19.7 \\
    Kyrgyz & 85.7 & 40.0 \\
    Tatar & 154.2 & 47.8 \\
    Turkish & 204.6 & 69.6 \\
    Uyghur & 180.1 & 51.1 \\
    Uzbek & 161.4 & 16.2 \\
    \hline
    \end{tabular}
    
    \caption{Average length of questions and answers across languages. An answer here refers to all choices, not only the correct ones.}
    \label{tab:stats}
\end{table}

\paragraph{Data composition} TUMLU contains eleven subjects: Maths, Physics, Chemistry, Biology, Geography, Native Language \& Literature (NL\&L), History, Logic, Human \& Society, Philosophy, Religion \& Ethics. Among these, Logic, Human \& Society, Philosophy and Religion \& Ethics subjects are available only in one or two languages. Therefore, they have not been included in experiments.

We report the number of characters per question and per choice in Table \ref{tab:stats}. High variance in question and answer length indicates variable question types and levels across languages. 

\paragraph{Difficulty levels}
While TUMLU can be used as a monolingual benchmark for any of the languages included, we are also interested in comparing performance across languages. This raises an important question: how comparable are questions of the same subject across different languages? While we can easily compare the model performance within each language, comparing it across languages proves more challenging. Different language subsets have different levels of difficulty. Uzbek and Turkish datasets are particularly difficult because those questions are designed specifically to imitate university entrance examinations in respective countries. Azerbaijani and Kazakh questions were developed by a community of students and teachers. While they all refer to middle- and high-school topics, there has been no oversight regarding their difficulty levels. For example, we know that maths questions in Kazakh are easier than the ones in other languages because they cover middle-school topics only. While these are explicit discrepancies that can be fixed in the future, there are certainly less obvious differences that are harder to identify and even harder to fix. That being said, even though comparison across languages is challenging, these datasets are more useful for comparison across models in monolingual settings.

\paragraph{TUMLU-mini} To make our experiments more balanced and less costly, we developed TUMLU-mini, which consists of 100 randomly selected and manually verified questions per subject. In cases where we had less than 100 questions, we used the entire set. You can find the number of questions per language and subject in Appendix \ref{sec:tumlu-mini}. If a question had more than 5 answer choices, one was dropped. All choices have been shuffled to make the dataset more robust to simple memorization since it is possible that these questions were a part of the pre-training corpus for these LLMs. We also removed subjects that were available in less than 3 languages. All experiments were run on this subset.

\section{Experimental set-up}
\label{section:setup}
\paragraph{Data} Previous work \cite{turkishmmlu} has shown that 100 questions per subject are enough to estimate the performance of a larger dataset. Therefore, we run all experiments on TUMLU-mini. While we have performed the experiments and publicly released the data, results on the following subjects are not reported in the paper: Logic, Philosophy, Religion \& Ethics, and Human \& Society. These subjects are available in one or two languages only, which makes any generalization impossible.

\paragraph{Model choice} We have used TUMLU to evaluate both open-source models, such as Llama 3.1 \cite{llama3}, Gemma 2 \cite{gemma2}, Qwen2.5 \cite{qwen25} and proprietary models, such as Gemini 1.5 \cite{gemini15}, Claude 3.5 \cite{claude35}, GPT-4o \cite{hurst2024gpt}. The size of selected open-source models varies between 7B and 70B. We do not have this information on proprietary models. This list includes models from the same series, such as Qwen2.5 7B instruct and Qwen2.5 70B instruct \cite{qwen25}, which allows us to observe the effect of scaling \cite{scaling2017, scaling2020} on multilingual performance. All open-source models are instruct-tuned versions. We have omitted this information in the tables to preserve space. Wherever applicable, we have included the performance of Claude 3.5 Sonnet in the paper, since it consistently outperforms all other models. The performance of the remaining models can be found in the appendices \ref{sec:5-shot-appendix} and \ref{sec:cot}.

\paragraph{Prompting}
We have run experiments in two settings: 5-shot, where we provide 5 example questions and answers on the same subject before asking a question \cite{fewshot}, and 5-shot CoT, where we provide 5 example questions and explanations of their answers before asking the question \cite{cot}. Few-shot and CoT prompt samples are available in Appendix \ref{sec:prompt-samples}. Previous work has demonstrated that \cite{include} providing the prompt in English does not result in performance gains. Due to this, we provide all prompts in respective native languages.

\paragraph{Technical details} We run our experiments through OpenAI API, Anthropic API, Google Cloud Gemini API, Together AI API, and Deep Infra API. No model was run on a local machine. We used the following hyperparameters with all APIs: TEMPERATURE = 0.0, MAX\_TOKENS = 1024, TOP\_P = 1.0.

\section{Results}
In this section, we present the few-shot and CoT performance of selected models on the TUMLU-mini dataset. We also present an analysis of output language. Lastly, we explore how well LLMs perform on the same questions written in different (Latin, Cyrillic, or Arabic) scripts.
\paragraph{5-shot results}
We present the average performance of all models in each language in Table \ref{tab:5-shot-all-models}. Claude 3.5 Sonnet outperforms other models in all languages. The top 5 spots belong to proprietary models, although it has to be noted that there are larger open-source models that have not been included in this benchmark. Among the available open-source models, Qwen2.5 72B Instruct has the best performance. Results also confirm the scaling hypothesis: Llama 3.1 70B significantly outperforms Llama 3.1 8B. The same applies to Qwen2.5 7B/72B and Gemma 2 9B/27B. We can also observe a significant improvement from Llama 3.1 70B to Llama 3.3 70B. While it is not possible to directly compare results across languages, we can observe that low-resource languages, such as Crimean Tatar, Karakalpak, and Uyghur have comparable performance to middle- and high-resource languages. Notably, this trend holds even with the lowest-performing models.

We present the 5-shot evaluation of Claude 3.5 Sonnet in more detail in Table \ref{tab:claude-3-5-sonnet-20241022_5shot}. In most languages, Native Language \& Literature is the most challenging subject for Claude 3.5 Sonnet. This holds for other models, as well (See Appendix \ref{sec:5shot}).

\begin{table*}
    \centering
    \begin{tabular}{lcccccccccc}
\hline
\textbf{Model} & \textbf{mean} & \textbf{aze} & \textbf{crh} & \textbf{kaa} & \textbf{kaz} & \textbf{kir} & \textbf{tat} & \textbf{tur} & \textbf{uig} & \textbf{uzb}\\
\hline
Claude 3.5 Sonnet & \textbf{78.9} & \textbf{84.4} & \textbf{81.2} & \textbf{75.3} & \textbf{83.0} & \textbf{75.7} & \textbf{84.0} & \textbf{85.7} & \textbf{71.3} & \textbf{69.1}\\ 
GPT-4o & 74.9 & 82.4 & 70.5 & 70.8 & 81.0 & 72.9 & 80.5 & 83.7 & 66.5 & 65.4\\ 
Gemini 1.5 Pro & 73.8 & 78.6 & 70.3 & 68.2 & 78.4 & 72.3 & 80.5 & 80.0 & 71.0 & 65.1\\ 
Gemini 1.5 Flash & 65.4 & 72.4 & 68.0 & 61.2 & 68.6 & 63.2 & 68.3 & 76.6 & 57.8 & 52.1\\ 
Claude 3.5 Haiku & 64.0 & 70.6 & 62.9 & 55.2 & 69.9 & 64.8 & 67.5 & 78.0 & 56.6 & 50.3\\ 
Llama- 3.1 405B & 63.0 & 65.9 & 69.5 & 60.0 & 69.0 & 64.1 & 70.4 & 59.7 & 58.2 & 50.4\\ 
Qwen2.5 72B & 59.9 & 70.1 & 61.8 & 54.6 & 62.6 & 47.5 & 62.5 & 73.9 & 56.0 & 50.4\\ 
Llama 3.3 70B & 58.6 & 66.0 & 58.7 & 49.2 & 60.0 & 60.2 & 69.5 & 68.4 & 51.6 & 44.1\\ 
Llama 3.1 70B & 57.7 & 68.1 & 57.3 & 49.9 & 56.4 & 58.4 & 66.2 & 64.9 & 52.4 & 45.3\\ 
Gemma 2 27b & 51.8 & 58.1 & 49.8 & 47.6 & 58.4 & 53.4 & 54.9 & 64.3 & 42.2 & 37.6\\ 
Gemma 2 9b & 47.1 & 53.7 & 46.8 & 40.8 & 49.1 & 48.9 & 51.8 & 60.4 & 35.8 & 36.1\\ 
Qwen2.5 7B & 41.0 & 48.0 & 42.6 & 37.2 & 45.0 & 31.8 & 40.5 & 55.6 & 33.4 & 34.6\\ 
Llama 3.1 8B  & 40.3 & 48.4 & 35.7 & 33.4 & 46.4 & 41.8 & 44.1 & 47.7 & 35.0 & 29.9\\ 
\hline
\end{tabular}
\caption{Average 5-shot performance of models on Azerbaijani (aze), Crimean Tatar (crh), Karakalpak (kaa), Kazakh (kaz), Kyrgyz (kir), Tatar (tat), Turkish (tur), Uyghur (uig), and Uzbek (uzb) datasets.}
\label{tab:5-shot-all-models}
\end{table*}

\begin{table*}[t]
\centering
\begin{tabular}{lccccccc}
\hline
\textbf{Language} & \textbf{Biology} & \textbf{Chemistry} & \textbf{Geography} & \textbf{History} & \textbf{Maths} & \textbf{NL\&L} & \textbf{Physics}\\
\hline
Azerbaijani & 89.0 & 89.0 & 91.0 & 71.0 & 85.0 & 73.0 & 93.0 \\
Crimean-tatar & 81.6 & 75.0 & 87.0 & 89.9 & 70.4 & 75.0 & 89.7 \\
Karakalpak & 78.0 & 85.7 & 75.0 & - & - & 42.2 & 95.6 \\
Kazakh & 92.0 & 73.0 & 78.0 & 78.0 & 96.0 & 76.0 & 88.0 \\
Kyrgyz & 77.0 & 80.0 & 87.0 & 78.6 & - & 80.8 & - \\
Tatar & 94.0 & 84.0 & 83.0 & 91.0 & 86.3 & 69.0 & 81.0 \\
Turkish & 84.0 & 88.0 & 94.0 & 92.0 & 78.0 & 79.0 & 85.0 \\
Uyghur & 75.0 & 66.0 & - & - & 75.8 & 66.0 & 73.5 \\
Uzbek & 71.0 & 73.0 & 64.0 & 70.0 & 70.0 & 55.0 & 81.0 \\
\hline
\end{tabular}
\caption{Subject-wise 5-shot performance of Claude 3.5 Sonnet across Turkic languages. Missing values indicate the absence of data for that language in the given subject. NL\&L refers to Native Language and Literature. $^\text{*}$This subset contains questions with 2 or 3 choices.}
\label{tab:claude-3-5-sonnet-20241022_5shot}
\end{table*}

\begin{table*}[t]
    \centering
    \begin{tabular}{lcccccc}

\hline
\textbf{Model} & \textbf{Mean} & \textbf{aze} & \textbf{kaz} & \textbf{tat} & \textbf{tur} & \textbf{uzb}\\ 
\hline
Claude 3.5 Sonnet & 84.0 & \textbf{87.1 \textcolor{Green}{(+2.7)}} & \textbf{84.1 \textcolor{Green}{(+1.1)}} & \textbf{87.9 \textcolor{Green}{(+3.9)}} & \textbf{87.9 \textcolor{Green}{(+2.1)}} & \textbf{72.9 \textcolor{Green}{(+3.7)}} \\
GPT-4o & 79.4 & 82.9 \textcolor{Green}{(+0.4)} & 80.7 \textcolor{Red}{(-0.3)} & 83.0 \textcolor{Green}{(+2.5)} & 84.0 \textcolor{Green}{(+0.3)} & 66.3 \textcolor{Green}{(+0.9)}\\
Gemini 1.5 Pro & 76.6 & 80.0 \textcolor{Green}{(+1.4)} & 75.1 \textcolor{Red}{(-3.3)} & 79.9 \textcolor{Red}{(-0.6)} & 81.0 \textcolor{Green}{(+1.0)} & 67.0 \textcolor{Green}{(+1.9)}\\ 			
Llama 3.1 405B & 68.9 & 73.4 \textcolor{Green}{(+7.6)}& 68.7 \textcolor{Red}{(-0.3)}& 65.0 \textcolor{Red}{(-5.4)}& 80.7 \textcolor{Green}{(+21.0)}& 56.4 \textcolor{Green}{(+6.0)}\\ 			
Claude 3.5 Haiku & 70.0 & 77.0 \textcolor{Green}{(+6.4)}& 74.0 \textcolor{Green}{(+4.1)}& 72.2 \textcolor{Green}{(+4.8)}& 77.6 \textcolor{Red}{(-0.4)}& 49.0 \textcolor{Red}{(-1.3)}\\ 
Gemini 1.5 Flash & 68.1 & 73.9 \textcolor{Green}{(+1.4)}& 69.0 \textcolor{Green}{(+0.4)}& 70.1 \textcolor{Green}{(+1.8)}& 73.6 \textcolor{Red}{(-3.0)}& 54.1 \textcolor{Green}{(+2.0)}\\ 			
Qwen2.5 72B & 67.1 & 72.1 \textcolor{Green}{(+2.0)}& 63.9 \textcolor{Green}{(+1.3)}& 67.6 \textcolor{Green}{(+5.1)}& 78.4 \textcolor{Green}{(+4.6)}& 53.6 \textcolor{Green}{(+3.1)}\\ 			
Llama 3.3 70B & 66.8 & 70.6 \textcolor{Green}{(+4.6)}& 69.3 \textcolor{Green}{(+9.3)}& 66.3 \textcolor{Red}{(-3.2)}& 77.4 \textcolor{Green}{(+9.0)}& 50.4 \textcolor{Green}{(+6.3)}\\ 	
Gemma 2 27B & 59.4 & 63.0 \textcolor{Green}{(+4.9)}& 61.6 \textcolor{Green}{(+3.1)}& 61.0 \textcolor{Green}{(+6.1)}& 66.4 \textcolor{Green}{(+2.1)}& 44.9 \textcolor{Green}{(+7.3)}\\ 			
Llama 3.1 70B & 56.2 & 59.4 \textcolor{Red}{(-8.7)}& 61.7 \textcolor{Green}{(+5.3)}& 58.6 \textcolor{Green}{(-7.6)}& 73.3 \textcolor{Green}{(+8.4)}& 27.9 \textcolor{Red}{(-17.4)}\\ 			
Gemma 2 9B & 52.0 & 57.3  \textcolor{Green}{(+3.6)}& 52.7  \textcolor{Green}{(+3.6)}& 50.1 \textcolor{Red}{(-1.7)}& 62.3  \textcolor{Green}{(+1.9)}& 37.4  \textcolor{Green}{(+1.3)}\\
Qwen2.5 7B & 46.4 & 48.1 \textcolor{Green}{(+0.1)}& 46.4 \textcolor{Green}{(+1.4)}& 43.0 \textcolor{Green}{(+2.5)}& 56.3 \textcolor{Green}{(+0.7)}& 38.0 \textcolor{Green}{(+3.4)}\\ 			
Llama 3.1 8B & 38.2 & 40.7 \textcolor{Red}{(-7.7)}& 38.9 \textcolor{Red}{(-7.6)}& 39.7 \textcolor{Red}{(-4.4)}& 45.1 \textcolor{Red}{(-2.6)}& 26.6 \textcolor{Red}{(-3.3)}\\ 			
\hline
\end{tabular}
\caption{Average 5-shot performance of models on Turkic languages.}
\label{tab:avg_5shot_cot}
\end{table*}

\begin{table*}[t]
\centering
\resizebox{1.9\columnwidth}{!}{%
\begin{tabular}{lccccccc}
\hline
\textbf{Language} & \textbf{Biology} & \textbf{Chemistry} & \textbf{Geography} & \textbf{History} & \textbf{Maths} & \textbf{NL\&L} & \textbf{Physics}\\
\hline
Azerbaijani & 89.0 \textcolor{Black}{(0.0)} & 96.0 \textcolor{Green}{(+7.0)} & 91.0 (0.0) & 78.0 \textcolor{Green}{(+7.0)} & 83.0 \textcolor{Red}{(-2.0)} & 76.0 \textcolor{Green}{(+3.0)} & 97.0 \textcolor{Green}{(+4.0)} \\
Kazakh & 96.0 \textcolor{Green}{(+4.0)} & 74.0 \textcolor{Green}{(+1.0)} & 78.0 (0.0) & 80.0 \textcolor{Green}{(+2.0)} & 95.0 \textcolor{Red}{(-1.0)} & 79.0 \textcolor{Green}{(+3.0)} & 87.0 \textcolor{Red}{(-1.0)} \\
Turkish & 86.0 \textcolor{Green}{(+2.0)} & 85.0 \textcolor{Red}{(-2.0)} & 95.0 \textcolor{Green}{(+1.0)} & 93.0 \textcolor{Green}{(+2.0)} & 77.0 (0.0) & 87.0 \textcolor{Green}{(+11.0)} & 89.0 \textcolor{Green}{(+4.0)} \\
Uzbek & 77.0 \textcolor{Green}{(+8.0)} & 73.0 (0.0) & 65.0 \textcolor{Green}{(+1.0)} & 65.0 \textcolor{Red}{(-3.0)} & 79.0 \textcolor{Green}{(+9.0)} & 45.0 \textcolor{Red}{(-10.0)} & 87.0 \textcolor{Green}{(+6.0)} \\
\hline
\end{tabular}
}
\caption{Subject-wise 5-shot CoT performance of Claude 3.5 Sonnet across Turkic languages.}
\label{tab:claude-3-5-sonnet-20241022_cot}
\end{table*}
\paragraph{5-shot CoT results}
We present the average results of the 5-shot CoT evaluation in Table \ref{tab:avg_5shot_cot}. CoT prompts have an overall positive effect on performance. Sporadic negative effects can be explained by incorrect output format, rather than incorrect answers. We avoided manual validation of the output and instead relied on generalizable pattern-matching methods.

Table \ref{tab:claude-3-5-sonnet-20241022_cot} shows the performance of Claude 3.5 Sonnet on each subject and language. On average, CoT prompts have a net positive effect in each subject and each language.

\begin{figure}
    \centering
    \includegraphics[width=\linewidth]{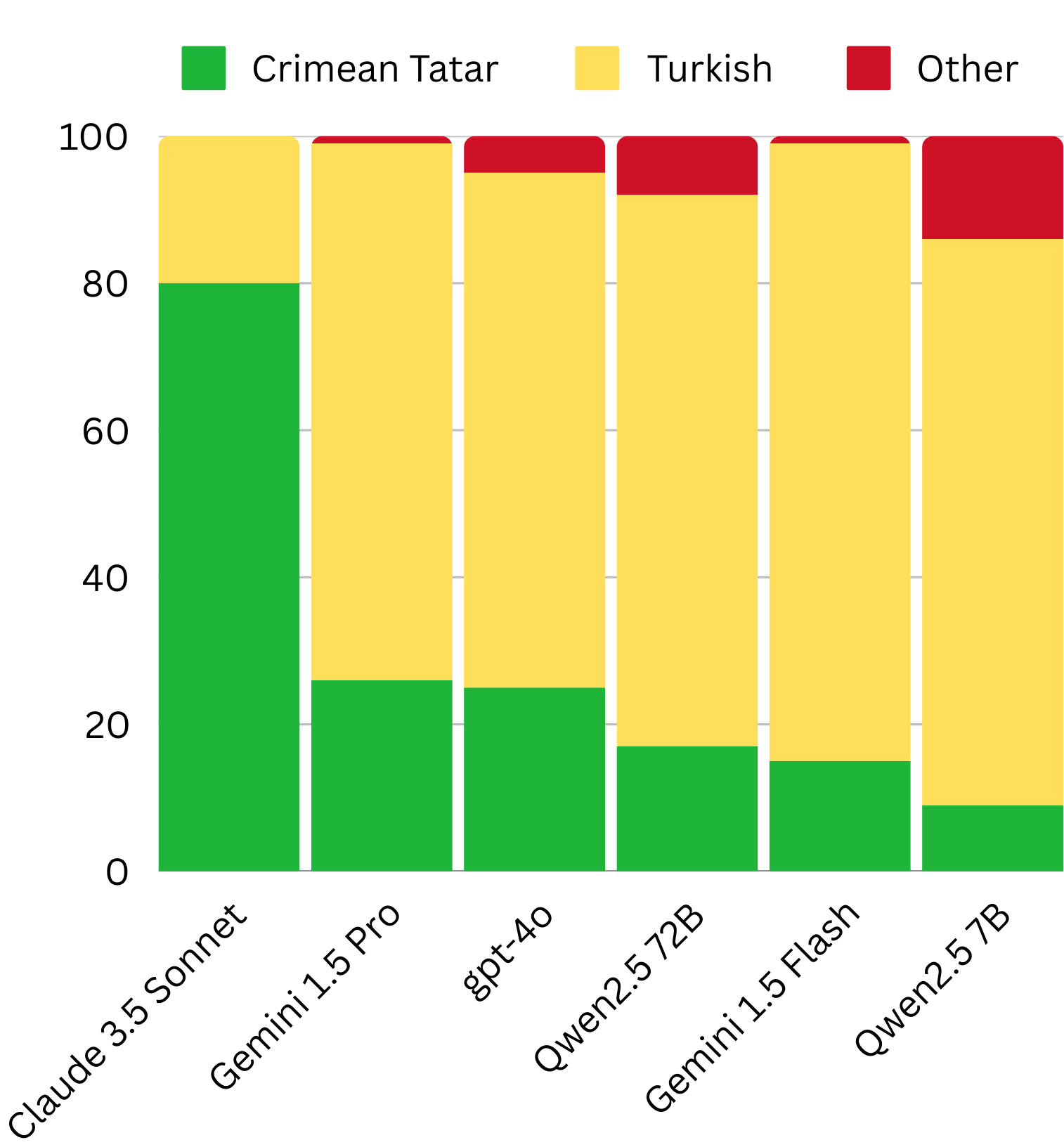}
    \caption{Language distribution of model responses to Crimean Tatar queries, as detected by Google Cloud Translation API.}
    \label{fig:crimean}
\end{figure}

\paragraph{Generated language vs. performance}
Benchmark results demonstrate that LLMs can have significant language understanding capabilities even in previously unseen languages, such as Crimean Tatar. This can be explained by linguistic proximity to languages better represented in the training data. Even though LLMs perform surprisingly well in these languages with simple 5-shot prompts. The results are less impressive when we analyze the generated text quality. While quality \textit{per se} is hard to quantify, we can detect the language of generated content. We used Google Cloud Translate API to detect output language. This API supports all languages in our benchmark, except for Karakalpak. We present results for Crimean Tatar in Figure \ref{fig:crimean}. As you can see, although these models have answered the majority of the questions in Crimean Tatar correctly, only a small portion of the generated text is classified as Crimean Tatar. Almost all of the answers are a synthesis between Turkish and Crimean Tatar. A similar issue appears in Kazakh when we switch from Cyrillic to Latin script. Although this has a small negative effect on the performance, the nature of the generated content changes dramatically. While the output of Cyrillic questions is easily detected as Kazakh, the output of Latin questions is easily confused with Tatar language.

\paragraph{Comparing performance on same questions written in different alphabets}
Some Turkic languages, such as Crimean Tatar, Kazakh, and Uzbek have both Cyrillic and Latin alphabets that are actively used. As a result, the text corpora that are used to train LLMs contain both versions. Also, transliteration between these scripts can be done automatically with a negligible error rate. Using these facts, we developed dual datasets for the languages above (see Figure \ref{fig:dual_dataset}). We evaluated models in both versions and compared their performance. We present some of the results in Table \ref{tab:scripts}. While the results initially seem irregular, they follow a simple pattern:
\begin{enumerate}
    \item In Crimean Tatar questions, all three models perform better in the Latin script. FineWeb 2 \cite{fineweb-2}, one of the largest multilingual text corpora, contains 21,365,608 Latin and 1,934,168 Cyrillic words in Crimean Tatar.
    \item In Kazakh questions, all three models perform better in the Cyrillic script. This aligns with the fact that most of the Kazakh text data on the web is written in the Cyrillic script. For example, the FineWeb 2 corpus contains 1,837,049,585 Cyrillic and 0 Latin words in Kazakh.
    \item In Uyghur questions, all three models perform better in the Arabic script. While Uyghur is not represented in Fineweb 2 corpus, virtually all Uyghur text is written in Arabic script.
    \item In Uzbek questions, results are less predictable. This can be explained by the fact that Cyrillic and Latin are more evenly distributed in Uzbek text. FineWeb 2 corpus contains 616,563,348 Latin and 492,264,125 Cyrillic words in Uzbek.\footnote{In this work, Uzbek refers to Northern Uzbek.}
\end{enumerate}
While these patterns hold across multiple models, there are exceptions. For example, on Uyghur questions, GPT-4o performs similarly with Arabic and Latin scripts. Llama 3.1 70B has an average accuracy of 28.48 on Uyghur questions with Arabic script and 41.30 with Latin script.

\begin{table*}
\small
\centering
\begin{tabular}{lccccccccc}
\hline
\textbf{Language} & \multicolumn{3}{c}{\textbf{Claude 3.5 Sonnet}} & \multicolumn{3}{c}{\textbf{Qwen2.5 72B}} & \multicolumn{3}{c}{\textbf{Gemma 2 27B}} \\ 
                  & \textbf{Cyrillic} & \textbf{Latin} & \textbf{Arabic} & \textbf{Cyrillic} & \textbf{Latin} & \textbf{Arabic} & \textbf{Cyrillic} & \textbf{Latin} & \textbf{Arabic} \\ \hline
Crimean Tatar    & 66.1 & \textbf{80.0} & —    & 47.6 & \textbf{61.8} & —    & 43.5 & \textbf{49.8} & —    \\
Kazakh           & \textbf{82.7} & 78.0 & —    & \textbf{64.3} & 54.1 & —    & \textbf{58.5} & 46.3 & —    \\
Uyghur           & —     & 64.5 & \textbf{70.8} & —     & 53.4 & \textbf{56.1} & —     & 36.0 & \textbf{42.2} \\ 
Uzbek            & 67.9 & \textbf{68.6} & —    & \textbf{51.1} & 50.4 & —    & \textbf{39.4} & 36.9 & —    \\ \hline
\end{tabular}

\caption{Performance comparison (\%) of three LLMs on Turkic languages with their native writing systems: Arabic and Latin for Uyghur, Cyrillic and Latin for Kazakh, Crimean Tatar, and Uzbek. Bold numbers indicate the best script performance per language-model pair. Dashes (—) denote script combinations not used in practice.}
\label{tab:scripts}
\end{table*}

\section{Conclusion}
We introduce TUMLU, a unified and native language understanding benchmark for Turkic languages. It contains 38139 multiple-choice questions in 8 languages and 11 subjects. Latin, Cyrillic, and Arabic scripts are represented in the benchmark. Uzbek, Crimean Tatar, and Kazakh are available in both Cyrillic and Latin. Uyghur is available both in Arabic and Latin. We also release TUMLU-mini, a smaller, more balanced and manually verified version that is more suitable for large-scale experiments. We use TUMLU-mini to benchmark 5 proprietary and 7 open-source LLMs. Results show that LLMs have a reasonably good understanding of all 8 languages, including ones that are not explicitly included in the training data of LLMs. However, LLMs are less capable of generating text in these languages, usually answering multiple-choice questions correctly, but in another, similar high-resource language. 
\section{Limitations}

TUMLU benchmark has two main limitations.

\paragraph{Mismatched difficulty levels} Native language \& literature subset contained both literature and language questions in some languages, while it contained only language questions in others. Similarly, the history subset contained both world and national history questions in some languages, while it contained only national questions in others. Maths questions in Kazakh are at middle-school level, which results in very high scores.

\paragraph{Missing major languages} TUMLU covers 8 Turkic languages with more than 180 million native speakers. However, some major Turkic languages, such as Turkmen, Kyrgyz and Bashkir are not included. We are hoping to extend our benchmark with more languages in further editions.

\section*{Acknowledgments}
We thank the Ukrainian Center for Educational Quality Assessment for providing exam materials in Crimean Tatar. We thank the Google Gemini Academic Program for supporting our experiments with cloud credits. We thank the Azerbaijan Cultural Society of Northern California for their financial support.

\bibliography{acl_latex}
\clearpage
\appendix
\section{Prompt samples}
\label{sec:prompt-samples}
\begin{figure*}
    \includegraphics[width=\linewidth]{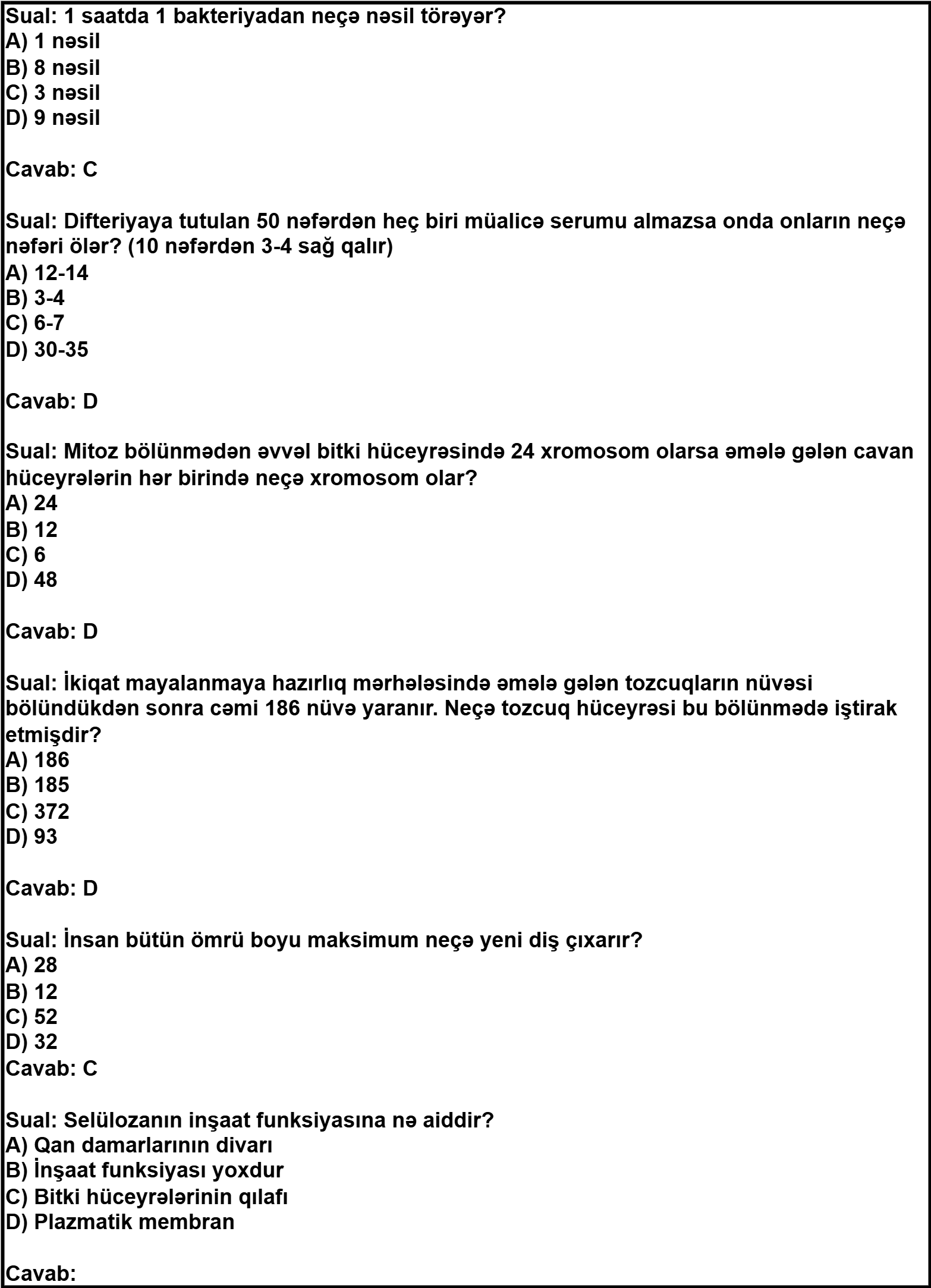}
    \caption{5-shot prompt sample for Biology questions in Azerbaijani.}
    \label{fig:5shot}
\end{figure*}

\begin{figure*}
    \includegraphics[width=\linewidth]{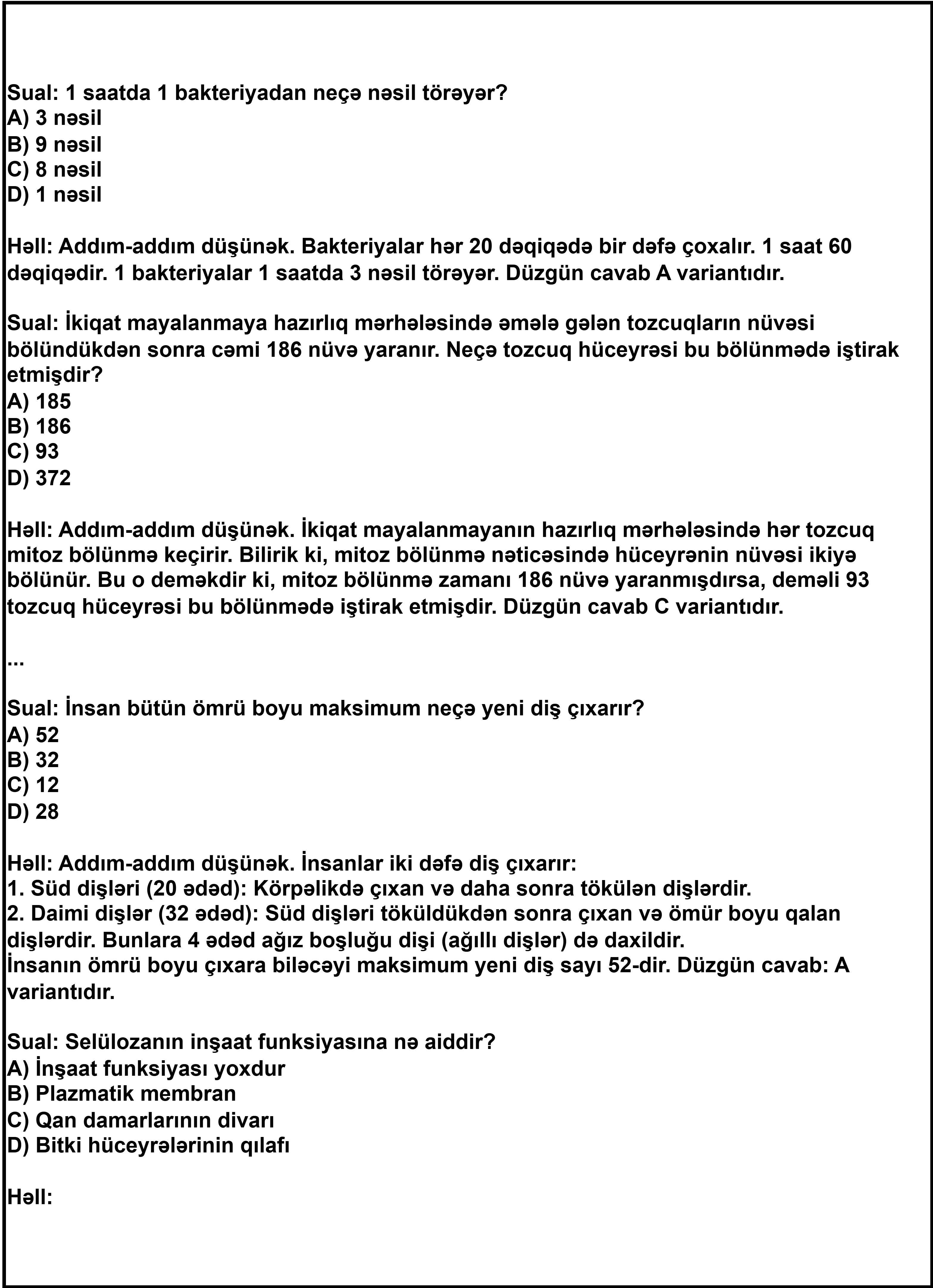}
    \caption{5-shot CoT prompt sample for Biology questions in Azerbaijani.}
    \label{fig:5shotcot}
\end{figure*}
\clearpage
\section{TUMLU-mini}
\label{sec:tumlu-mini}
\begin{table*}[h]
\begin{tabular}{lrrrrrrr}
\hline
\textbf{Language (code)}      & \multicolumn{1}{l}{\textbf{NL\&L}} & \multicolumn{1}{l}{\textbf{History}} & \multicolumn{1}{l}{\textbf{Geography}} & \multicolumn{1}{l}{\textbf{Chemistry}} & \multicolumn{1}{l}{\textbf{Physics}} & \multicolumn{1}{l}{\textbf{Biology}} & \multicolumn{1}{l}{\textbf{Maths}} \\
\hline
Azerbaijani (aze)   & 100                             & 100                         & 100                           & 100                           & 100                         & 100                         & 100                       \\
Crimean Tatar (crh) & 100                             & {\color[HTML]{F8A102} 69}   & {\color[HTML]{F8A102} 23}     & {\color[HTML]{F8A102} 32}     & {\color[HTML]{F8A102} 39}   & {\color[HTML]{F8A102} 38}   & {\color[HTML]{F8A102} 54} \\
Karakalpak (kaa)    & {\color[HTML]{F8A102} 64}       & {\color[HTML]{FE0000} 0}    & {\color[HTML]{F8A102} 28}     & {\color[HTML]{F8A102} 28}     & {\color[HTML]{F8A102} 45}   & {\color[HTML]{F8A102} 50}   & {\color[HTML]{FE0000} 0}  \\
Kazakh (kaz)        & 100                             & 100                         & 100                           & 100                           & 100                         & 100                         & 100                       \\
Kyrgyz (kir)        & 100                             & 98                         & 100                           & 100                           & 100                         & 100                         & 100                       \\
Tatar (tat)         & 100                             & 100                         & 100                           & 100                           & 100                         & 100                         & 100                       \\
Turkish (tur)       & 100                             & 100                         & 100                           & 100                           & 100                         & 100                         & 100                       \\
Uyghur (uig)        & 100                             & {\color[HTML]{FE0000} 0}    & {\color[HTML]{FE0000} 0}      & {\color[HTML]{F8A102} 97}     & {\color[HTML]{F8A102} 98}   & 100                         & {\color[HTML]{F8A102} 99} \\
Uzbek (uzb)         & 100                             & 100                         & 100                           & 100                           & 100                         & 100                         & 100         \\
\hline
\end{tabular}
\label{tab:tumlu-mini}
\caption{Composition of TUMLU-mini, a more balanced and manually verified subset of TUMLU benchmark. All experiments in this paper have been run on TUMLU-mini. These numbers exclude sample questions used in 5-shot and 5-shot CoT prompts. Language codes are from ISO 639-3.}
\end{table*}
\clearpage
\section{5-shot results}
\label{sec:5shot}
This appendix includes 5-shot results for all models, except for Claude 3.5 Sonnet which is available in Table \ref{tab:claude-3-5-sonnet-20241022_5shot}.

\begin{table*}[h]
\centering
\begin{tabular}{lccccccc}
\hline
\textbf{Language} & \textbf{Biology} & \textbf{Chemistry} & \textbf{Geography} & \textbf{History} & \textbf{Maths} & \textbf{NL\&L} & \textbf{Physics}\\
\hline
Azerbaijani & 78.00 & 79.00 & 72.00 & 61.00 & 65.00 & 58.00 & 81.00 \\
Crimean Tatar & 63.16 & 53.12 & 73.91 & 56.52 & 57.41 & 62.00 & 65.52 \\
Karakalpak & 60.00 & 60.71 & 53.57 & - & - & 20.31 & 80.00 \\
Kazakh & 79.00 & 66.00 & 72.00 & 64.00 & 76.00 & 58.00 & 74.00 \\
Tatar & 74.00 & 63.00 & 73.00 & 74.00 & 63.16 & 59.00 & 65.00 \\
Turkish & 71.00 & 82.00 & 84.00 & 85.00 & 71.00 & 70.00 & 80.00 \\
Uyghur & 57.00 & 44.33 & - & - & 59.60 & 52.00 & 59.18 \\
Uzbek & 53.00 & 50.00 & 48.00 & 49.00 & 54.00 & 41.00 & 55.00 \\
\hline
\end{tabular}
\caption{Accuracy scores for Claude 3.5 Haiku-20241022 model across languages.}
\end{table*}
\label{sec:5-shot-appendix}

\begin{table*}
\centering
\begin{tabular}{lccccccc}
\hline
\textbf{Language} & \textbf{Biology} & \textbf{Chemistry} & \textbf{Geography} & \textbf{History} & \textbf{Maths} & \textbf{NL\&L} & \textbf{Physics}\\
\hline
Azerbaijani & 80.00 & 80.00 & 78.00 & 55.00 & 72.00 & 56.00 & 84.00 \\
Crimean Tatar & 78.95 & 59.38 & 73.91 & 59.42 & 61.11 & 67.00 & 72.41 \\
Karakalpak & 70.00 & 82.14 & 53.57 & - & - & 31.25 & 68.89 \\
Kazakh & 88.00 & 60.00 & 73.00 & 57.00 & 68.00 & 57.00 & 76.00 \\
Tatar & 86.00 & 68.00 & 74.00 & 68.00 & 64.21 & 47.00 & 60.00 \\
Turkish & 75.00 & 72.00 & 78.00 & 76.00 & 78.00 & 59.00 & 74.00 \\
Uyghur & 71.00 & 53.61 & - & - & 59.60 & 57.00 & 45.92 \\
Uzbek & 59.00 & 43.00 & 49.00 & 57.00 & 69.00 & 22.00 & 51.00 \\
\hline
\end{tabular}
\caption{Accuracy scores for GEMINI-1.5-FLASH model across languages.}
\end{table*}

\begin{table*}
\centering
\begin{tabular}{lccccccc}
\hline
\textbf{Language} & \textbf{Biology} & \textbf{Chemistry} & \textbf{Geography} & \textbf{History} & \textbf{Maths} & \textbf{NL\&L} & \textbf{Physics}\\
\hline
Azerbaijani & 81.00 & 86.00 & 76.00 & 56.00 & 78.00 & 57.00 & 89.00 \\
Crimean Tatar & 71.05 & 50.00 & 65.22 & 56.52 & 61.11 & 52.00 & 62.07 \\
Karakalpak & 76.00 & 71.43 & 57.14 & - & - & 43.75 & 88.89 \\
Kazakh & 88.00 & 68.00 & 75.00 & 73.00 & 94.00 & 70.00 & 80.00 \\
Tatar & 95.00 & 78.00 & 81.00 & 84.00 & 80.00 & 59.00 & 64.00 \\
Turkish & 51.00 & 61.00 & 61.00 & 70.00 & 64.00 & 50.00 & 57.00 \\
Uyghur & 70.00 & 42.27 & - & - & 49.49 & 66.00 & 51.02 \\
Uzbek & 59.00 & 69.00 & 61.00 & 54.00 & 79.00 & 30.00 & 76.00 \\
\hline
\end{tabular}
\caption{Accuracy scores for GEMINI-1.5-PRO model across languages.}
\end{table*}

\begin{table*}
\centering
\begin{tabular}{lccccccc}
\hline
\textbf{Language} & \textbf{Biology} & \textbf{Chemistry} & \textbf{Geography} & \textbf{History} & \textbf{Maths} & \textbf{NL\&L} & \textbf{Physics}\\
\hline
Azerbaijani & 61.00 & 57.00 & 59.00 & 47.00 & 45.00 & 42.00 & 65.00 \\
Crimean Tatar & 50.00 & 37.50 & 56.52 & 52.17 & 33.33 & 53.00 & 44.83 \\
Karakalpak & 48.00 & 46.43 & 35.71 & - & - & 25.00 & 48.89 \\
Kazakh & 67.00 & 37.00 & 63.00 & 41.00 & 38.00 & 48.00 & 50.00 \\
Tatar & 69.00 & 53.00 & 63.00 & 54.00 & 35.79 & 41.00 & 47.00 \\
Turkish & 65.00 & 55.00 & 76.00 & 75.00 & 45.00 & 48.00 & 57.00 \\
Uyghur & 40.00 & 26.80 & - & - & 37.37 & 38.00 & 36.73 \\
Uzbek & 49.00 & 33.00 & 39.00 & 41.00 & 31.00 & 26.00 & 34.00 \\
\hline
\end{tabular}
\caption{Accuracy scores for GOOGLE/GEMMA-2-9B-IT model across languages.}
\end{table*}

\begin{table*}
\centering
\begin{tabular}{lccccccc}
\hline
\textbf{Language} & \textbf{Biology} & \textbf{Chemistry} & \textbf{Geography} & \textbf{History} & \textbf{Maths} & \textbf{NL\&L} & \textbf{Physics}\\
\hline
Azerbaijani & 70.00 & 59.00 & 62.00 & 46.00 & 38.00 & 55.00 & 77.00 \\
Crimean Tatar & 44.74 & 46.88 & 60.87 & 49.28 & 35.19 & 60.00 & 51.72 \\
Karakalpak & 52.00 & 64.29 & 42.86 & - & - & 23.44 & 55.56 \\
Kazakh & 79.00 & 44.00 & 65.00 & 56.00 & 52.00 & 51.00 & 62.00 \\
Tatar & 71.00 & 58.00 & 71.00 & 63.00 & 43.16 & 32.00 & 45.00 \\
Turkish & 73.00 & 65.00 & 81.00 & 78.00 & 41.00 & 55.00 & 57.00 \\
Uyghur & 50.00 & 35.05 & - & - & 34.34 & 51.00 & 40.82 \\
Uzbek & 46.00 & 27.00 & 40.00 & 44.00 & 35.00 & 26.00 & 40.00 \\
\hline
\end{tabular}
\caption{Accuracy scores for GOOGLE/GEMMA-2-27B-IT model across languages.}
\end{table*}

\begin{table*}
\centering
\begin{tabular}{lccccccc}
\hline
\textbf{Language} & \textbf{Biology} & \textbf{Chemistry} & \textbf{Geography} & \textbf{History} & \textbf{Maths} & \textbf{NL\&L} & \textbf{Physics}\\
\hline
Azerbaijani & 91.00 & 93.00 & 89.00 & 75.00 & 70.00 & 67.00 & 92.00 \\
Crimean Tatar & 60.53 & 59.38 & 69.57 & 86.96 & 57.41 & 70.00 & 89.66 \\
Karakalpak & 80.00 & 82.14 & 71.43 & - & - & 35.94 & 84.44 \\
Kazakh & 93.00 & 71.00 & 76.00 & 77.00 & 85.00 & 77.00 & 88.00 \\
Tatar & 98.00 & 78.00 & 88.00 & 92.00 & 69.47 & 69.00 & 68.00 \\
Turkish & 86.00 & 79.00 & 95.00 & 94.00 & 63.00 & 82.00 & 87.00 \\
Uyghur & 84.00 & 54.64 & - & - & 62.63 & 65.00 & 66.33 \\
Uzbek & 70.00 & 68.00 & 65.00 & 69.00 & 56.00 & 51.00 & 79.00 \\
\hline
\end{tabular}
\caption{Accuracy scores for GPT-4o model across languages.}
\end{table*}

\begin{table*}
\centering
\begin{tabular}{lccccccc}
\hline
\textbf{Language} & \textbf{Biology} & \textbf{Chemistry} & \textbf{Geography} & \textbf{History} & \textbf{Maths} & \textbf{NL\&L} & \textbf{Physics}\\
\hline
Azerbaijani & 76.00 & 72.00 & 77.00 & 59.00 & 45.00 & 49.00 & 84.00 \\
Crimean Tatar & 68.42 & 53.12 & 69.57 & 72.46 & 40.74 & 58.00 & 48.28 \\
Karakalpak & 58.00 & 46.43 & 64.29 & - & - & 21.88 & 55.56 \\
Kazakh & 80.00 & 42.00 & 71.00 & 62.00 & 52.00 & 51.00 & 60.00 \\
Tatar & 88.00 & 67.00 & 83.00 & 82.00 & 67.37 & 51.00 & 48.00 \\
Turkish & 76.00 & 58.00 & 86.00 & 83.00 & 41.00 & 65.00 & 68.00 \\
Uyghur & 37.00 & 46.39 & - & - & 48.48 & 49.00 & 46.94 \\
Uzbek & 52.00 & 38.00 & 52.00 & 50.00 & 44.00 & 31.00 & 42.00 \\
\hline
\end{tabular}
\caption{Accuracy scores for META-LLAMA/LLAMA-3.3-70B-INSTRUCT model across languages.}
\end{table*}

\begin{table*}
\centering
\begin{tabular}{lccccccc}
\hline
\textbf{Language} & \textbf{Biology} & \textbf{Chemistry} & \textbf{Geography} & \textbf{History} & \textbf{Maths} & \textbf{NL\&L} & \textbf{Physics}\\
\hline
Azerbaijani & 53.00 & 50.00 & 57.00 & 45.00 & 36.00 & 47.00 & 51.00 \\
Crimean Tatar & 42.11 & 34.38 & 30.43 & 44.93 & 18.52 & 45.00 & 34.48 \\
Karakalpak & 38.00 & 39.29 & 32.14 & - & - & 21.88 & 35.56 \\
Kazakh & 60.00 & 33.00 & 64.00 & 54.00 & 32.00 & 41.00 & 41.00 \\
Tatar & 55.00 & 40.00 & 61.00 & 55.00 & 33.68 & 33.00 & 31.00 \\
Turkish & 51.00 & 37.00 & 63.00 & 50.00 & 34.00 & 35.00 & 41.00 \\
Uyghur & 38.00 & 27.84 & - & - & 36.36 & 42.00 & 30.61 \\
Uzbek & 31.00 & 22.00 & 38.00 & 33.00 & 25.00 & 29.00 & 31.00 \\
\hline
\end{tabular}
\caption{Accuracy scores for META-LLAMA/META-LLAMA-3.1-8B-INSTRUCT model across languages.}
\end{table*}

\begin{table*}
\centering
\begin{tabular}{lccccccc}
\hline
\textbf{Language} & \textbf{Biology} & \textbf{Chemistry} & \textbf{Geography} & \textbf{History} & \textbf{Maths} & \textbf{NL\&L} & \textbf{Physics}\\
\hline
Azerbaijani & 78.00 & 67.00 & 78.00 & 60.00 & 48.00 & 62.00 & 80.00 \\
Crimean Tatar & 63.16 & 40.62 & 65.22 & 66.67 & 29.63 & 62.00 & 65.52 \\
Karakalpak & 58.00 & 53.57 & 60.71 & - & - & 28.12 & 44.44 \\
Kazakh & 57.00 & 27.00 & 55.00 & 57.00 & 40.00 & 50.00 & 48.00 \\
Tatar & 72.00 & 23.00 & 67.00 & 72.00 & 49.47 & 52.00 & 47.00 \\
Turkish & 70.00 & 55.00 & 88.00 & 70.00 & 40.00 & 62.00 & 68.00 \\
Uyghur & 24.00 & 16.49 & - & - & 20.20 & 45.00 & 36.73 \\
Uzbek & 42.00 & 39.00 & 53.00 & 34.00 & 36.00 & 27.00 & 42.00 \\
\hline
\end{tabular}
\caption{Accuracy scores for META-LLAMA/META-LLAMA-3.1-70B-INSTRUCT model across languages.}
\end{table*}

\begin{table*}
\centering
\begin{tabular}{lccccccc}
\hline
\textbf{Language} & \textbf{Biology} & \textbf{Chemistry} & \textbf{Geography} & \textbf{History} & \textbf{Maths} & \textbf{NL\&L} & \textbf{Physics}\\
\hline
Azerbaijani & 41.00 & 58.00 & 53.00 & 37.00 & 59.00 & 30.00 & 58.00 \\
Crimean Tatar & 36.84 & 37.50 & 39.13 & 39.13 & 55.56 & 42.00 & 48.28 \\
Karakalpak & 30.00 & 42.86 & 39.29 & - & - & 25.00 & 48.89 \\
Kazakh & 38.00 & 44.00 & 54.00 & 31.00 & 64.00 & 31.00 & 53.00 \\
Tatar & 41.00 & 38.00 & 44.00 & 42.00 & 56.84 & 28.00 & 34.00 \\
Turkish & 42.00 & 59.00 & 62.00 & 69.00 & 58.00 & 40.00 & 59.00 \\
Uyghur & 34.00 & 29.90 & - & - & 38.38 & 40.00 & 24.49 \\
Uzbek & 35.00 & 31.00 & 30.00 & 34.00 & 52.00 & 21.00 & 39.00 \\
\hline
\end{tabular}
\caption{Accuracy scores for QWEN/QWEN2.5-7B-INSTRUCT model across languages.}
\end{table*}

\begin{table*}
\centering
\begin{tabular}{lccccccc}
\hline
\textbf{Language} & \textbf{Biology} & \textbf{Chemistry} & \textbf{Geography} & \textbf{History} & \textbf{Maths} & \textbf{NL\&L} & \textbf{Physics}\\
\hline
Azerbaijani & 76.00 & 77.00 & 74.00 & 53.00 & 73.00 & 54.00 & 84.00 \\
Crimean Tatar & 65.79 & 53.12 & 60.87 & 72.46 & 55.56 & 66.00 & 58.62 \\
Karakalpak & 50.00 & 75.00 & 50.00 & - & - & 35.94 & 62.22 \\
Kazakh & 60.00 & 55.00 & 64.00 & 52.00 & 75.00 & 52.00 & 79.00 \\
Tatar & 68.00 & 60.00 & 65.00 & 78.00 & 71.58 & 41.00 & 54.00 \\
Turkish & 79.00 & 73.00 & 84.00 & 85.00 & 61.00 & 56.00 & 79.00 \\
Uyghur & 60.00 & 51.55 & - & - & 64.65 & 52.00 & 52.04 \\
Uzbek & 53.00 & 49.00 & 44.00 & 55.00 & 63.00 & 28.00 & 61.00 \\
\hline
\end{tabular}
\caption{Accuracy scores for QWEN/Qwen2.5 72B-INSTRUCT model across languages.}
\end{table*}
\clearpage
\section{5-shot CoT results}
\label{sec:cot}
This appendix includes 5-shot CoT results for all models, except for Claude 3.5 Sonnet which is available in Table \ref{tab:claude-3-5-sonnet-20241022_cot}.

\begin{table*}[h]
\centering
\begin{tabular}{lccccccc}
\hline
\textbf{Language} & \textbf{Biology} & \textbf{Chemistry} & \textbf{Geography} & \textbf{History} & \textbf{Maths} & \textbf{NL\&L} & \textbf{Physics}\\
\hline
Azerbaijani & 81.00 & 83.00 & 81.00 & 69.00 & 76.00 & 61.00 & 88.00 \\
Kazakh & 82.00 & 65.00 & 75.00 & 61.00 & 87.00 & 64.00 & 84.00 \\
Turkish & 74.00 & 78.00 & 91.00 & 85.00 & 28.00 & 71.00 & 68.00 \\
Uzbek & 50.00 & 50.00 & 37.00 & 47.00 & 63.00 & 24.00 & 53.00 \\
\hline
\end{tabular}
\caption{Accuracy scores for Claude 3.5 Haiku-20241022 model across languages.}
\end{table*}

\begin{table*}
\centering
\begin{tabular}{lccccccc}
\hline
\textbf{Language} & \textbf{Biology} & \textbf{Chemistry} & \textbf{Geography} & \textbf{History} & \textbf{Maths} & \textbf{NL\&L} & \textbf{Physics}\\
\hline
Azerbaijani & 81.00 & 83.00 & 81.00 & 52.00 & 71.00 & 59.00 & 88.00 \\
Kazakh & 79.00 & 61.00 & 69.00 & 51.00 & 85.00 & 50.00 & 78.00 \\
Turkish & 71.00 & 67.00 & 69.00 & 69.00 & 76.00 & 60.00 & 74.00 \\
Uzbek & 50.00 & 56.00 & 41.00 & 49.00 & 70.00 & 17.00 & 67.00 \\
\hline
\end{tabular}
\caption{Accuracy scores for GEMINI-1.5-FLASH model across languages.}
\end{table*}

\begin{table*}
\centering
\begin{tabular}{lccccccc}
\hline
\textbf{Language} & \textbf{Biology} & \textbf{Chemistry} & \textbf{Geography} & \textbf{History} & \textbf{Maths} & \textbf{NL\&L} & \textbf{Physics}\\
\hline
Azerbaijani & 79.00 & 92.00 & 78.00 & 53.00 & 78.00 & 62.00 & 80.00 \\
Kazakh & 83.00 & 70.00 & 63.00 & 65.00 & 92.00 & 67.00 & 83.00 \\
Turkish & 73.00 & 76.00 & 84.00 & 79.00 & 85.00 & 51.00 & 77.00 \\
Uzbek & 53.00 & 67.00 & 43.00 & 40.00 & 73.00 & 17.00 & 81.00 \\
\hline
\end{tabular}
\caption{Accuracy scores for GEMINI-1.5-PRO model across languages.}
\end{table*}

\begin{table*}
\centering
\begin{tabular}{lccccccc}
\hline
\textbf{Language} & \textbf{Biology} & \textbf{Chemistry} & \textbf{Geography} & \textbf{History} & \textbf{Maths} & \textbf{NL\&L} & \textbf{Physics}\\
\hline
Azerbaijani & 64.00 & 67.00 & 65.00 & 39.00 & 45.00 & 47.00 & 73.00 \\
Kazakh & 61.00 & 42.00 & 64.00 & 46.00 & 61.00 & 32.00 & 62.00 \\
Turkish & 72.00 & 60.00 & 74.00 & 71.00 & 45.00 & 50.00 & 64.00 \\
Uzbek & 41.00 & 31.00 & 40.00 & 41.00 & 26.00 & 25.00 & 32.00 \\
\hline
\end{tabular}
\caption{Accuracy scores for GOOGLE/GEMMA-2-9B-IT model across languages.}
\end{table*}

\begin{table*}
\centering
\begin{tabular}{lccccccc}
\hline
\textbf{Language} & \textbf{Biology} & \textbf{Chemistry} & \textbf{Geography} & \textbf{History} & \textbf{Maths} & \textbf{NL\&L} & \textbf{Physics}\\
\hline
Azerbaijani & 72.00 & 76.00 & 71.00 & 41.00 & 55.00 & 48.00 & 77.00 \\
Kazakh & 74.00 & 52.00 & 63.00 & 46.00 & 70.00 & 53.00 & 73.00 \\
Turkish & 77.00 & 64.00 & 80.00 & 77.00 & 53.00 & 49.00 & 65.00 \\
Uzbek & 43.00 & 43.00 & 48.00 & 53.00 & 42.00 & 10.00 & 49.00 \\
\hline
\end{tabular}
\caption{Accuracy scores for GOOGLE/GEMMA-2-27B-IT model across languages.}
\end{table*}

\begin{table*}
\centering
\begin{tabular}{lccccccc}
\hline
\textbf{Language} & \textbf{Biology} & \textbf{Chemistry} & \textbf{Geography} & \textbf{History} & \textbf{Maths} & \textbf{NL\&L} & \textbf{Physics}\\
\hline
Azerbaijani & 90.00 & 91.00 & 88.00 & 73.00 & 75.00 & 70.00 & 93.00 \\
Kazakh & 89.00 & 63.00 & 78.00 & 82.00 & 85.00 & 84.00 & 84.00 \\
Turkish & 86.00 & 80.00 & 97.00 & 92.00 & 70.00 & 80.00 & 83.00 \\
Uzbek & 73.00 & 68.00 & 70.00 & 74.00 & 59.00 & 39.00 & 79.00 \\
\hline
\end{tabular}
\caption{Accuracy scores for GPT-4o model across languages.}
\end{table*}

\begin{table*}
\centering
\begin{tabular}{lccccccc}
\hline
\textbf{Language} & \textbf{Biology} & \textbf{Chemistry} & \textbf{Geography} & \textbf{History} & \textbf{Maths} & \textbf{NL\&L} & \textbf{Physics}\\
\hline
Azerbaijani & 81.00 & 78.00 & 75.00 & 54.00 & 70.00 & 52.00 & 79.00 \\
Kazakh & 80.00 & 57.00 & 70.00 & 65.00 & 74.00 & 59.00 & 80.00 \\
Turkish & 74.00 & 67.00 & 84.00 & 87.00 & 77.00 & 67.00 & 75.00 \\
Uzbek & 46.00 & 15.00 & 8.00 & 33.00 & 35.00 & 10.00 & 14.00 \\
\hline
\end{tabular}
\caption{Accuracy scores for META-LLAMA/LLAMA-3.3-70B-INSTRUCT model across languages.}
\end{table*}

\begin{table*}
\centering
\begin{tabular}{lccccccc}
\hline
\textbf{Language} & \textbf{Biology} & \textbf{Chemistry} & \textbf{Geography} & \textbf{History} & \textbf{Maths} & \textbf{NL\&L} & \textbf{Physics}\\
\hline
Azerbaijani & 43.00 & 44.00 & 57.00 & 36.00 & 20.00 & 32.00 & 53.00 \\
Kazakh & 52.00 & 28.00 & 55.00 & 46.00 & 17.00 & 32.00 & 42.00 \\
Turkish & 50.00 & 37.00 & 58.00 & 61.00 & 20.00 & 39.00 & 50.00 \\
Uzbek & 34.00 & 5.00 & 37.00 & 29.00 & 9.00 & 23.00 & 9.00 \\
\hline
\end{tabular}
\caption{Accuracy scores for META-LLAMA/META-LLAMA-3.1-8B-INSTRUCT model across languages.}
\end{table*}

\begin{table*}
\centering
\begin{tabular}{lccccccc}
\hline
\textbf{Language} & \textbf{Biology} & \textbf{Chemistry} & \textbf{Geography} & \textbf{History} & \textbf{Maths} & \textbf{NL\&L} & \textbf{Physics}\\
\hline
Azerbaijani & 52.00 & 67.00 & 69.00 & 49.00 & 55.00 & 50.00 & 71.00 \\
Kazakh & 77.00 & 51.00 & 65.00 & 65.00 & 56.00 & 53.00 & 65.00 \\
Turkish & 72.00 & 64.00 & 83.00 & 88.00 & 69.00 & 60.00 & 77.00 \\
Uzbek & 31.00 & 24.00 & 8.00 & 20.00 & 12.00 & 11.00 & 15.00 \\
\hline
\end{tabular}
\caption{Accuracy scores for META-LLAMA/META-LLAMA-3.1-70B-INSTRUCT model across languages.}
\end{table*}

\begin{table*}
\centering
\begin{tabular}{lccccccc}
\hline
\textbf{Language} & \textbf{Biology} & \textbf{Chemistry} & \textbf{Geography} & \textbf{History} & \textbf{Maths} & \textbf{NL\&L} & \textbf{Physics}\\
\hline
Azerbaijani & 42.00 & 54.00 & 54.00 & 37.00 & 51.00 & 35.00 & 63.00 \\
Kazakh & 38.00 & 41.00 & 48.00 & 31.00 & 66.00 & 33.00 & 68.00 \\
Turkish & 51.00 & 59.00 & 65.00 & 48.00 & 70.00 & 43.00 & 58.00 \\
Uzbek & 40.00 & 30.00 & 40.00 & 42.00 & 52.00 & 17.00 & 44.00 \\
\hline
\end{tabular}
\caption{Accuracy scores for QWEN/QWEN2.5-7B-INSTRUCT model across languages.}
\end{table*}

\begin{table*}
\centering
\begin{tabular}{lccccccc}
\hline
\textbf{Language} & \textbf{Biology} & \textbf{Chemistry} & \textbf{Geography} & \textbf{History} & \textbf{Maths} & \textbf{NL\&L} & \textbf{Physics}\\
\hline
Azerbaijani & 72.00 & 88.00 & 80.00 & 51.00 & 80.00 & 46.00 & 88.00 \\
Kazakh & 64.00 & 50.00 & 70.00 & 50.00 & 84.00 & 52.00 & 77.00 \\
Turkish & 78.00 & 78.00 & 89.00 & 84.00 & 79.00 & 57.00 & 84.00 \\
Uzbek & 54.00 & 55.00 & 46.00 & 50.00 & 49.00 & 27.00 & 68.00 \\
\hline
\end{tabular}
\caption{Accuracy scores for QWEN/Qwen2.5 72B-INSTRUCT model across languages.}
\end{table*}
\end{document}